\documentclass[letterpaper]{article} 
\usepackage{aaai24}  
\usepackage{times}  
\usepackage{helvet}  
\usepackage{courier}  
\usepackage[hyphens]{url}  
\usepackage{graphicx} 
\usepackage{natbib}  
\usepackage{caption} 
\usepackage{algorithm}
\usepackage{algorithmic}

\usepackage{amssymb}
\usepackage{amsmath}
\usepackage{booktabs}
\usepackage{multirow}
\usepackage{newfloat}
\usepackage{listings}
\DeclareCaptionStyle{ruled}{labelfont=normalfont,labelsep=colon,strut=off} 
\floatstyle{ruled}
\newfloat{listing}{tb}{lst}{}
\floatname{listing}{Listing}
\title{What Makes Quantization for Large Language Models Hard?\\An Empirical Study from the Lens of Perturbation}
\author {
    Zhuocheng Gong\textsuperscript{\rm 1},
    Jiahao Liu\textsuperscript{\rm 2},
    Jingang Wang\textsuperscript{\rm 2},
    Xunliang Cai\textsuperscript{\rm 2},
    Dongyan Zhao\textsuperscript{\rm 1,4}\footnotemark[2],
    Rui Yan\textsuperscript{\rm 3}\footnotemark[2]
}
\affiliations {
    \textsuperscript{\rm 1} Wangxuan Institute of Computer Technology, Peking University \\
    \textsuperscript{\rm 2} Meituan\\
    \textsuperscript{\rm 3} Gaoling School of Artificial Intelligence, Renmin University of China \\
    \textsuperscript{\rm 4} State Key Laboratory of Media Convergence Production Technology and Systems \\
    \{gzhch,zhaody\}@pku.edu.cn, ruiyan@ruc.edu.cn, \{liujiahao12,wangjingang02,caixunliang\}@meituan.com
}

\usepackage{bibentry}

\begin{document}

\maketitle
\begin{abstract}
Quantization has emerged as a promising technique for improving the memory and computational efficiency of large language models (LLMs). Though the trade-off between performance and efficiency is well-known, there is still much to be learned about the relationship between quantization and LLM performance. To shed light on this relationship, we propose a new perspective on quantization, viewing it as perturbations added to the weights and activations of LLMs. We call this approach ``the lens of perturbation". Using this lens, we conduct experiments with various artificial perturbations to explore their impact on LLM performance. Our findings reveal several connections between the properties of perturbations and LLM performance, providing insights into the failure cases of uniform quantization and suggesting potential solutions to improve the robustness of LLM quantization.
To demonstrate the significance of our findings, we implement a simple non-uniform quantization approach based on our insights. Our experiments show that this approach achieves minimal performance degradation on both 4-bit weight quantization and 8-bit quantization for weights and activations. These results validate the correctness of our approach and highlight its potential to improve the efficiency of LLMs without sacrificing performance.
\end{abstract}

\renewcommand{\thefootnote}{\fnsymbol{footnote}}
\footnotetext[2]{Corresponding authors: Dongyan Zhao and Rui Yan.}

\section{Introduction}

Recently, large language models (LLMs) have gained a lot of attention in the deep learning community due to their remarkable performance~\cite{touvron2023llama,chatgpt}. However, their size and computational demands pose challenges for deployment on resource-constrained devices or in real-time applications. To address this issue and improve the accessibility and efficiency of these models, researchers have explored quantization techniques that enable high-precision data to be represented in lower-precision formats. These techniques can significantly reduce the memory footprint and computational requirements of large models, making them more practical for deployment and use.

\begin{figure}[h]
  \centering
  \includegraphics[width=\linewidth]{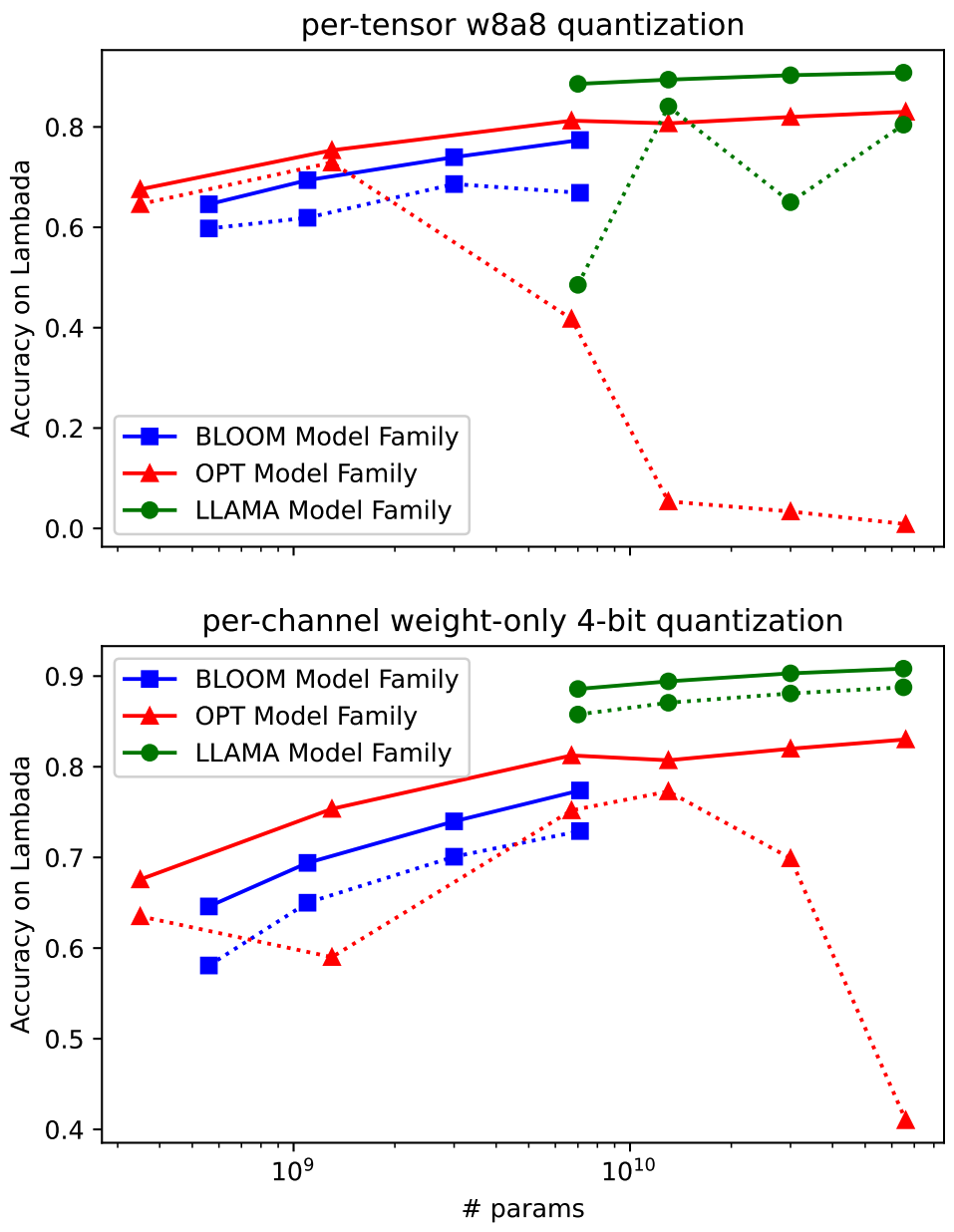}
  \caption{Performance of quantized LLMs on Lambada~\cite{paperno-EtAl:2016:P16-1} on different model families (BLOOM~\cite{scao2022bloom}, OPT~\cite{zhang2022opt}, and LLAMA~\cite{touvron2023llama}), the parameter number scaling from 350M to 66B. We implement two uniform quantization settings: one is to quantize both weights and activations to 8 bits (W8A8), and the other is to perform 4-bit channel-wise for weights only (W4A16).}
  \label{fig:1}
\end{figure}

Quantizing large language models presents unique challenges that require careful consideration. First, LLMs can be incredibly large, with billions of parameters and multiple layers, resulting in significant quantization error accumulation. Second, due to the large model scale, training can be extremely costly, making post-training quantization (PTQ) the most appealing approach for LLMs, as it requires much less or even zero calibration data to perform~\cite{bondarenko2021understanding,wu2023zeroquant,xiao2023smoothquant}.
In summary, effectively quantifying LLMs requires a tailored approach that balances model complexity, computational resources, and performance. Despite recent studies investigating the quantization of LLMs, a comprehensive investigation into the performance of quantization has not been satisfactorily undertaken. Specifically, exploring the impact of different quantization strategies on different model families during the quantization process would be beneficial to the development of LLM quantization. 

In this paper, our goal is to investigate the impact of quantization on LLMs in a systematic manner, with a focus on understanding the challenges of quantizing large language models. Specifically, we examine the effects of \textbf{zero-shot post-training uniform quantization} on LLMs, as it is a practical and representative quantization setting that requires no additional training or data.
To begin, we apply standard uniform quantization to various LLMs and analyze the resulting performance degradation (as shown in Figure~\ref{fig:1}). Our findings indicate that the impact of quantization varies significantly across different model families and sizes. The BLOOM and LLAMA model families exhibit greater robustness to quantization, while OPT models are more sensitive to quantization. Furthermore, we observe that the scale of the model is a crucial factor that affects the quality of quantization. Finally, we find that activation quantization has a significant effect on performance.

To account for the various factors that impact quantization, it is important to establish an analytical framework. This is where \textbf{"the lens of perturbation"} comes to play. We view quantization as the addition of small perturbations to the weights and/or activations of the model. By perturbing these components and measuring the resulting performance degradation, we can analyze the impact of quantization on LLMs. This approach enables us to gain insights into how quantization influences the model's performance in different ways.
To be more specific, we subject LLMs to different kinds of artificial perturbations and investigate which perturbation causes the most performance degradation. By experimenting with artificial perturbations, we can gain insights into the robustness of the LLM quantization method and identify potential areas for improvement.

By using the lens of perturbation, we are able to delve deeper into the mechanism of quantization and identify several factors that impact the performance of LLM quantization. Our first finding is that the ability of data to withstand perturbations is closely tied to its magnitude. Specifically, larger values can endure more significant perturbations with minimal performance degradation, while smaller values can only tolerate minor perturbations.
Our second observation pertains to the outlier phenomenon, which involves abnormally large values in weights and activations~\cite{bondarenko2021understanding}. These insights obtained through the lens of perturbation explain why uniform quantization may not always be optimal. Uniform quantization treats all weights and activations equally, without considering their sensitivity to perturbations, and is obsessed with outlier values. Instead, we propose a potential improvement by taking into account the sensitivity to perturbations.
Through experiments, we demonstrate that this modification significantly reduces the quantization error, where uniform quantization falls short. These findings provide a more nuanced understanding of the impact of quantization on LLMs and offer a promising direction for improving the quantization process.

In summary, our contributions are as follows:
\begin{itemize}
    \item We propose a new perspective of investigating uniform quantization for LLMs, namely the lens of perturbation. This lens provides new insights into the challenges of quantization for LLMs.
    \item We conduct a comprehensive comparison of the quantization performance of LLMs across various model scales and three model families (LLAMA, BLOOM, and OPT), using three quantization settings (W4A16, W8A8, and W4A8)
    \item Drawing on the insights through the lens of perturbation, we conduct preliminary experiments to leverage the properties of quantization-friendly perturbations and explore the potential of non-uniform quantization for LLMs, which significantly reduces quantization error. These experiments demonstrate that non-uniform quantization can significantly reduce quantization error.
\end{itemize}

\begin{figure*}[t]
  \centering
  \includegraphics[width=0.85\textwidth]{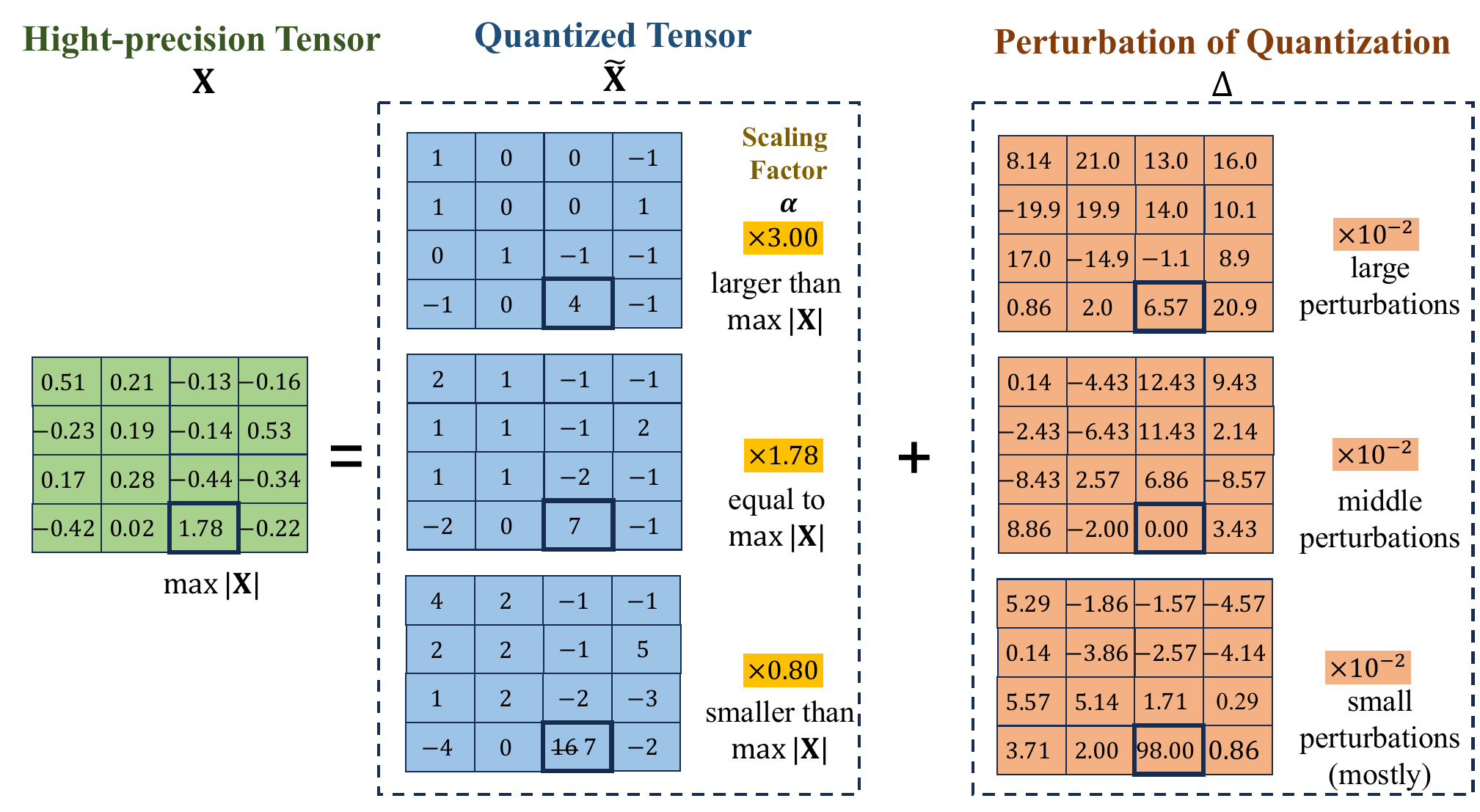}
  \caption{A toy example of 4-bit uniform quantization with different scale factors $\alpha$. The top line shows a choice of $\alpha$ that is too large, leading to too much information lost in the quantization process. The middle represents the most common choice, where $\alpha$ is set to the maximum value of the tensor. The bottom line uses a smaller $\alpha$, which results in less perturbation at the cost of clipping out-of-range values.}
  \label{fig:0}
\end{figure*}

\section{Related Works}
\subsection{Large Language Model Quantization}
Quantization, which represents the weights and activations of neural networks with low-bit precision, has been widely studied in computer vision and natural language processing (NLP) communities~\cite{gholami2021survey}.
Recently, some researchers explore quantizing large language models to reduce the deployment cost~\cite{shen2020q,wu2022extreme,kim2021bert,bondarenko2021understanding,wu2023zeroquant,liu2023llm,dettmers2023qlora,xiao2023smoothquant}. One major factor distinguishing LLM quantization is that LLMs with their enormous parameters are too expensive to train. Thus, most current approaches choose few-shot or zero-shot post-training quantization, which directly quantizes the already-trained model without much calibration data.
These methods mostly build on uniform quantization, the most widespread, practical, and easy-to-use quantization strategy. 
Up to now, it has been validated that quantizing the weight of LLMs to 8 or even 4 bits is not too hard~\cite{dettmers2023case}, however, activation quantization is still a major obstacle. Multiple studies have revealed the outlier phenomenon in weights and activations of transformer-based models, a tiny fixed set of embedding dimensions consistently exhibits large values across multiple layers~\cite{bondarenko2021understanding,puccetti2022outlier}. Statistical analysis of LLMs shows that activations suffer from outliers several orders of magnitude larger than weights~\cite{xiao2023smoothquant}. As a result, under the setting of uniform quantization, activations encounter much larger quantization loss. Some methods have been proposed to alleviate this problem, for example, by separating the quantization of outlier dimensions and regular dimensions~\cite{gong-etal-2023-prequant}, or by partially migrating outlier effect from activations to weights~\cite{xiao2023smoothquant}
. Recently, some researchers have analyzed why outliers appear and proposed a method to remove outliers and produce quantization-friendly models~\cite{bondarenko2023quantizable}.
While most existing works target proposing new quantization methods for LLMs, some focus on empirical analysis of LLM quantization, among which \citet{yao2023zeroquantv2} thoroughly compares the effect of different quantization schemes, model families, and quantization coverage, e.g. 

\section{Preliminaries}
\subsection{Uniform Quantization} 
Quantization is a technique that involves mapping high-precision values into low-precision discrete values. This process reduces the memory and computational requirements of the model, making it more efficient to deploy on devices with limited resources. Given a float-point tensor $\textbf{X}$ (either a weight tensor or activation tensor), the quantization and de-quantization process can be viewed as:
\begin{equation}
\label{eq:1}
    \textbf{X}^{\mathbb{Z}}= Q(\textbf{X}),\;\;\; \widetilde{\textbf{X}}=Q^{-1}(\textbf{X}^{\mathbb{Z}})
\end{equation}
where $\textbf{X}^{\mathbb{Z}}$ is the quantized tensor, and $\widetilde{\textbf{X}}$ is the recovered tensor.
In this paper, we target the setting of uniform quantization~\cite{hubara2017quantized}, which is the most common quantization because of its efficient implementation and satisfactory performance. The quantization function $Q(\cdot)$ of uniform quantization is defined as a rounding-to-nearest operation over the scaled input:
\begin{equation}
\label{eq:2}
    Q_{\text{uni}}(\textbf{X})= \text{clip}\left(\left\lfloor\frac{\textbf{X}}{\alpha}\cdot2^b\right\rceil+z;0,2^b-1\right)
\end{equation}
where $b\in\mathbb{N}$ is the bit-width, $\alpha\in\mathbb{R}$ is the scale factor, and $z\in\mathbb{N}$ is zero-point. $\lfloor\cdot\rceil$ denotes the round-to-nearest-integer operator.
The corresponding de-quantization function is as follows:
\begin{equation}
\label{eq:3}
    Q_{\text{uni}}^{-1}({\textbf{X}})=(\textbf{X}^{\mathbb{Z}}-z)\cdot\alpha/2^b
\end{equation}

In addition to the quantization strategy, the quality of quantization is also influenced by the bit-width and quantization granularity. The former is straightforward to comprehend: the number of quantization bins increases exponentially with the bit-width. For example, 8-bit quantization has 256 unique bins for each $\textbf{X}$. The quantization granularity determines the size of $\textbf{X}$. Tensor-wise quantization involves quantizing the entire tensor. Channel-wise or group-wise quantization, on the other hand, divides the tensor along a specific dimension and then quantizes each channel or group independently. Fine-grained granularity typically leads to better performance.

\subsection{The Lens of Perturbation}
While the quantized low-precision values are expected to be close to their high-precision counterparts, they usually may not be exactly equal due to the loss of precision. The extent of deviation from the high-precision values is influenced by the bit-width and the quantization scheme used.
One way to view quantization is as perturbations added to the original values, which can be expressed as:
\begin{equation}
\label{eq:4}
    \textbf{X}=\widetilde{\textbf{X}}+\Delta,
\end{equation}
where $\Delta$ is the difference matrix between the quantized and original tensors, representing the noise introduced by the quantization process. We call this perspective as \textbf{the lens of perturbation}. 

\begin{figure}[t]
  \centering
  \includegraphics[width=0.3\textwidth]{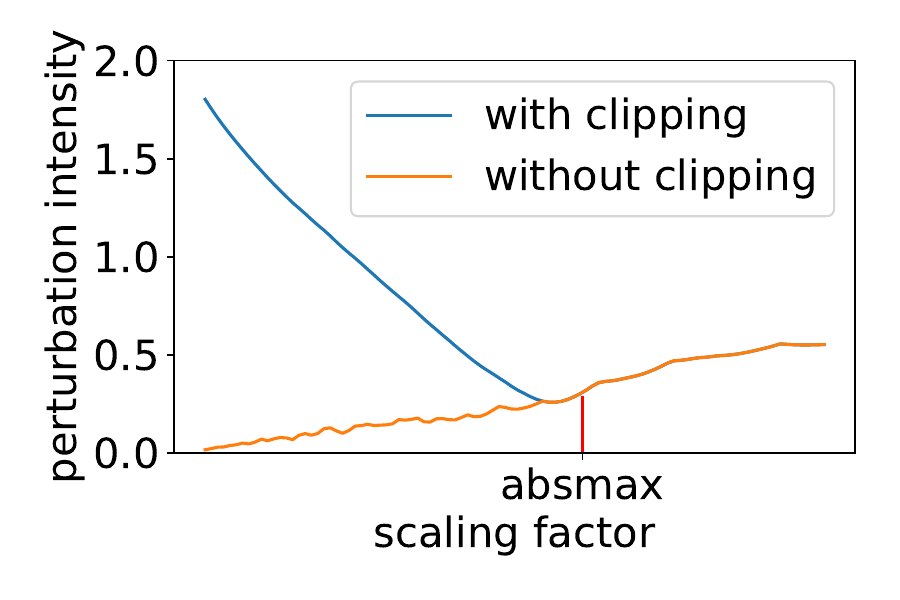}
  \caption{An illustration of the relationship between the scale factor and the intensity of perturbation. To calculate the perturbation intensity, we use the L2 norm of $\Delta$, which represents the distance between the high-precision tensor and the quantized one.}
  \label{fig:01}
\end{figure}

\subsection{The Choice of the Scale Factor}
In uniform quantization, the scale factor $\alpha$ plays a critical role as it determines the scaling of original values. Usually, the scale factor is computed by optimizing the performance of a few calibration samples. However, in the zero-shot setting, it is often set to the absolute maximum value, represented $\alpha_{\text{max}}=\max(|\textbf{X}|)$. 

Figure~\ref{fig:0} illustrates how the choice of scale factor impacts uniform quantization. If the scale factor is greater than $\max(|\textbf{X}|)$, values are overscaled, resulting in an insufficient utilization of the range of integers. Conversely, if the scale factor is smaller than $\max(|\textbf{X}|)$, quantized values exceed the range so out-of-range values would be clipped. 
In the example depicted in Figure~\ref{fig:0}, the maximum value of the input tensor 
Smaller scale factors result in smaller quantization intervals and thus slighter perturbations for in-range values. Figure~\ref{fig:01} provides a rough relationship between the intensity of perturbation and the scale factor. Throughout this paper, we use $\alpha_{\text{max}}$ as the scale factor for uniform quantization, unless otherwise specified.

\paragraph{Uniform Quantization under the Lens of Perturbation}
Before conducting experiments, we first examine uniform quantization through the lens of perturbation. For simplicity, we initially disregard the clipping operation for out-of-range values and later discuss how it affects quantization. In uniform quantization, high-precision values are mapped to a series of equal interval bins.  Through the lens of perturbation, we make the following observations: (1) the intensity of perturbation is positively correlated with the scale factor (as depicted in Figure~\ref{fig:01}, the larger the scale factor, the greater the intensity of perturbation); (2) due to the uniform intervals, the magnitude of perturbation is not dependent on the magnitude of the values being quantized. 

\section{Looking into LLM Quantization through the Lens of Perturbation}
In this section, we investigate the relationships between various types of perturbations and their impact on performance degradation. Analyzing the statistical properties of perturbation can provide a more profound comprehension of the effects of quantization. Our goal is to determine which type of perturbation is more quantization-friendly (leading to less performance degradation). 

\begin{figure*}[htbp]
  \centering
  \includegraphics[width=\textwidth]{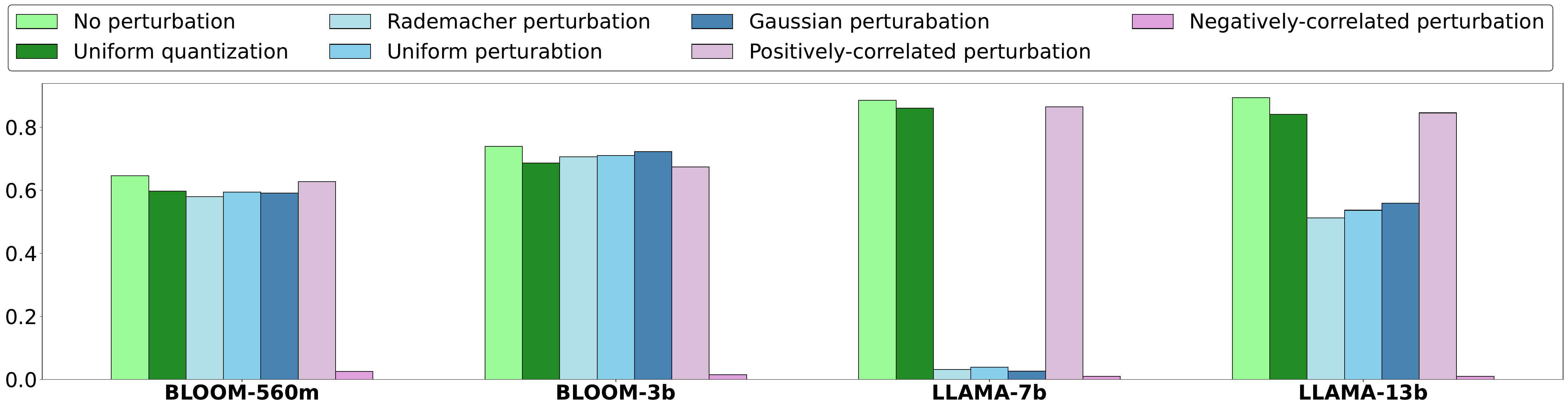}
  \caption{Lambada~\cite{paperno-EtAl:2016:P16-1} performance comparison on different types of perturbations. Here \textit{No perturbation} represents the vanilla LLM (without quantization and without perturbation). \textit{Uniform quantization} represents for W8A8 uniform quantization. For a fair comparison, all artificial perturbations are set to have the same variance with the native perturbation caused by \textit{Uniform quantization}. The perturbation methods are run four times with different random seeds.}
  \label{fig:2}
\end{figure*}

\subsection{What Kind of Perturbation Is More Quantization-Friendly?}
To explore the correlation between perturbations and their impact on performance, we propose to use manually constructed perturbations instead of relying on the natural perturbations caused by quantization, which can be intricate and inconvenient to manipulate. Our approach involves controlling the properties of artificial perturbations and examining how they influence the model's behavior.

Our approach constructs perturbations from two perspectives: the distribution of the perturbation (i.e., the type of distribution from which the perturbation is sampled) and the magnitude of the perturbation (i.e., whether the perturbation's size is proportional to the input's size). By manipulating these properties, we can enhance our comprehension of how perturbations impact the model's performance.
We consider three types of distributions for our perturbations:
\begin{itemize}
    \item \textbf{Gaussian Perturbation} $\Delta_{\text{G}} \sim \mathcal{N}(\mu,\sigma^2)$: where the perturbations are sampled from a Gaussian distribution.
    \item \textbf{Uniform Perturbation} $\Delta_{\text{U}} \sim \mathcal{U}(-c,c)$: where the perturbations follow the uniform distribution between $-c$ and $c$. 
    \item \textbf{Rademacher Perturbation} $\Delta_{\text{R}} \sim \mathcal{R}$: where the perturbations have constant sizes but random signs.
\end{itemize}

Apart from the distribution, we also take into account whether the magnitude of perturbations should be linked to the magnitude of the input (i.e., whether larger values should have greater (or lesser) perturbations). In this sense, the three methods mentioned above are classified as ``magnitude-independent perturbation," signifying that the perturbation's magnitude is not contingent on the input's magnitude
To investigate the impact of perturbation magnitude, we introduce two ``magnitude-aware perturbation" settings, in which the perturbation's size is dynamically determined by the value's size.
\begin{itemize}
    \item \textbf{Positively-correlated Perturbation} $\Delta_{\text{M+}}$: the positive correlation setting scales the perturbation magnitude positively with the input, meaning that higher weights have larger perturbations, formally, $\delta\propto|\textbf{x}|$. 
    \item \textbf{Negatively-correlated Perturbation} $\Delta_{\text{M-}}$: conversely, the negative correlation setting scales the perturbation magnitude inversely with the input, meaning that higher weights lead to smaller perturbations, formally, we choose $\delta\propto(|\textbf{x}|+\epsilon)^{-1}$.
\end{itemize}

We conducted experiments using different types of perturbations on the BLOOM-560M, BLOOM-3B, LLAMA-7B, and LLAMA-13B models. Due to the high computational cost, we did not experiment with more LLMs. Nevertheless, we are confident that our current experiments provide sufficient observations to draw reliable conclusions.
To simulate the W8A8 quantization setting, we add perturbations to both weights and activations in this experiment. To ensure a fair comparison between artificial perturbations and quantization perturbations, we control the perturbation intensity (measured by the L2 norm of $\Delta$) to be the same for each matrix being quantized.

\subsubsection{Observations}
The results of perturbations are shown in Figure~\ref{fig:2}. As expected, perturbations lead to non-negligible decreases in performance, regardless of the properties of the perturbations or the backbone models. The extent of performance degradation varies across different model families. In the case of the BLOOM model family, artificial perturbations exhibit similar performance to standard uniform quantization, whereas LLAMA models experience a significant loss in performance. This contrast highlights the distinctive characteristics of different LLM model families, which warrants further investigation.
Regarding our primary objective of establishing a link between the properties of perturbations and performance degradation, we have made an interesting discovery. We have found that the sampling distribution has little impact on performance, as all three distributions we tested exhibit comparable performance. On the other hand, the magnitude of perturbations plays a crucial role. Specifically, the Positively-correlated perturbation $\Delta_{\text{M+}}$ (where larger values have larger perturbations) exhibits a dominant advantage over all perturbations on all models, while the negatively-correlated perturbation $\Delta_{\text{M+}}$ (where smaller values have larger perturbations) almost completely destroys performance. This phenomenon has led us to a significant finding: \textbf{larger values in LLM activations and weights are more robust to perturbations, whereas smaller values are more sensitive to perturbations.}

\subsection{Clipping as a Special Perturbation}
\begin{figure}[tbp]
  \centering
  \includegraphics[width=0.48\textwidth]{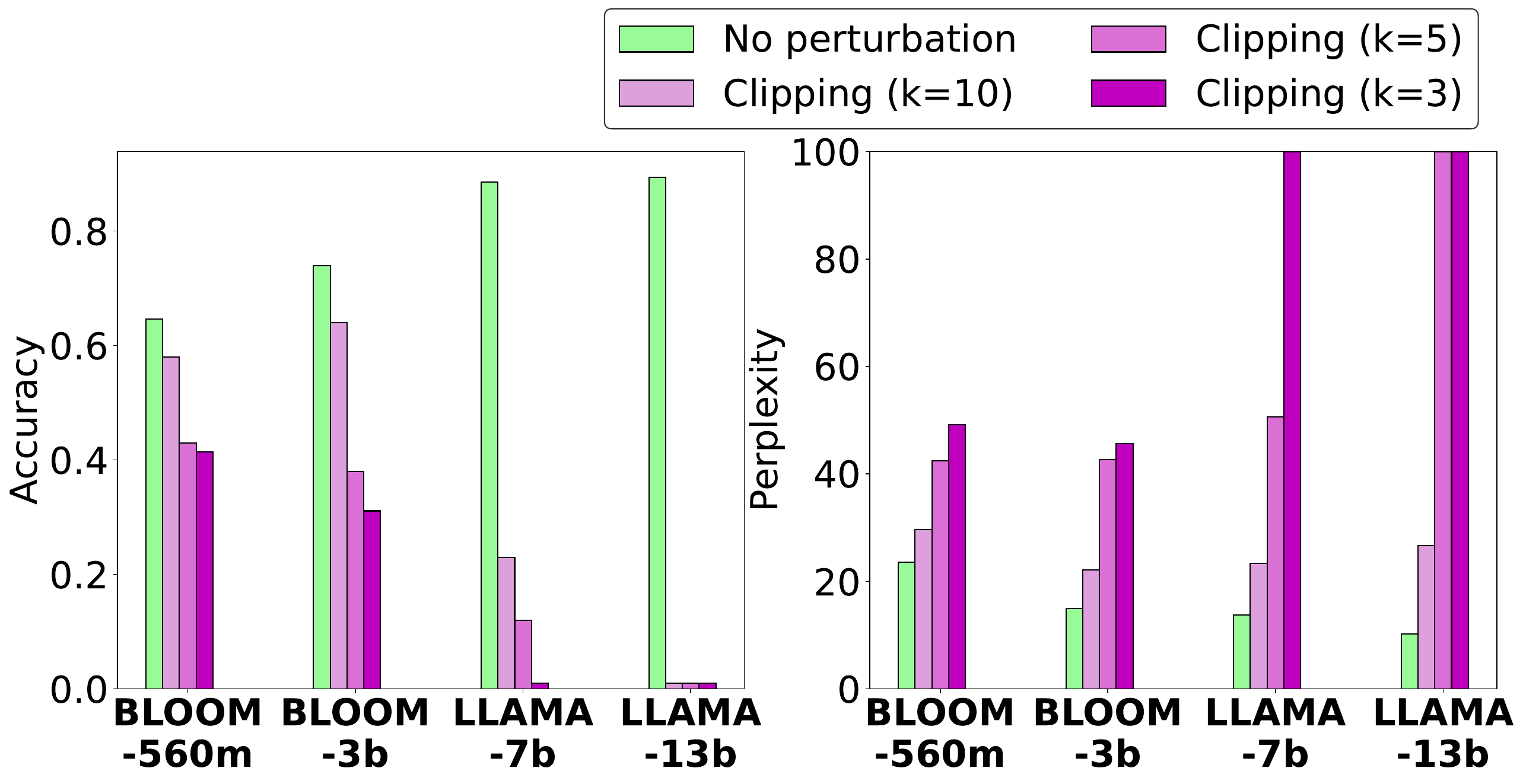}
  \caption{Effects of clipping on Lambada dataset (left) and Wikitext-2~\cite{merity2016pointer} and C4~\cite{2019t5} (averaged ppl.) (right). We compare three options of $k$, which determines the clipping threshold: $k\in\{3,5,10\}$. It is worth noting that even with the most stringent clipping setting, where $k=3$, on average, only less than 0.1\% of values are affected by clipping.}
  \label{fig:3}
\end{figure}

The experiment conducted on artificial perturbations mentioned above has not considered the effect of the clipping operation, which can have a considerable impact on realistic quantization scenarios, however.
According to the formula for uniform quantization (Eq.~\ref{eq:2}), if $\alpha$ is smaller than $\max(|\textbf{X}|)$, out-of-range values will be clipped, and the deviation caused by the clipping operation is much greater than that caused by regular perturbations. 
Through the lens of perturbation, clipping can be viewed as an exceedingly large perturbation applied to out-of-range values. Our current finding shows that larger values can tolerate much greater perturbations. 
Therefore, the question arises as to whether a perturbation as large as clipping is acceptable in terms of model performance.

\begin{table*}[t]
\small
\setlength\tabcolsep{4.pt}
\begin{center}
\scalebox{1}{\begin{tabular}{llrrrrrrrrrrrrrr}
\toprule
Quant. & Bits & \multicolumn{4}{c|}{Bloom} & \multicolumn{6}{|c|}{OPT} & \multicolumn{4}{|c}{Llama} \\
Methods & (W/A) & 560M & 1.1B & 3B & 7.1B & 350M & 1.3B & 6.7B & 13B & 30B & 66B  & 7B & 13B & 30B & 65B\\ \midrule
\multicolumn{2}{l}{\textit{Wiketext-2 (ppl$\downarrow$)}}
 \\ 
\midrule
full & 16/16 & 19.36 & 24.33 & 11.83 & 10.32 & 15.89 & 11.35 & 8.84 & 8.32 & 7.93 & 7.14  & 19.7 & 13.3 & 10.3 & 10.18 \\ \midrule
Uniform & 4/16 & 24.19 & 29.97 & 13.41 & 11.45 & 15.95 & 11.34 & 8.86 & 8.30 & 7.93 & 8.65 & 25.05 & 14.98 & 11.78 & 11.42 \\
Non uniform & 4/16 & 22.02 & 27.67 & 12.87 & 10.98 & 15.89 & 11.35 & 8.84 & 8.30 & 7.93 & 7.14 & 20.92 & 15.55 & 11.03 & 11.29 \\ \midrule
Uniform & 8/8 & 26.14 & 36.88 & 14.98 & 15.64 & 18.28 & 12.2 & 18.83 & 3300 & 1328 & 2254 & 100.81 & 52.41 & 57.91 & 26.88 \\
Non uniform & 8/8 & 19.78 & 24.42 & 11.90 & 10.36 & 15.95 & 11.51 & 8.86 & 8.34 & 7.93 & 7.38 & 22.28 & 16.14 & 12.62 & 15.13\\ \midrule
Uniform & 4/8 & 33.31 & 44.56 & 17.28 & 17.52 & 18.33 & 12.16 & 18.58 & 3352 & 1262 & 2379 & 453.05 & 75.81 & 711 & 1095 \\
Non uniform & 4/8 & 22.67 & 27.72 & 12.91 & 11.03 & 15.92 & 11.49 & 8.86 & 8.32 & 7.93 & 7.38 & 21.02 & 15.58 & 11.07 & 12.01 \\ \midrule
\multicolumn{2}{l}{\textit{C4 (ppl$\downarrow$)}}
\\ 
\midrule
full & 16/16 & 27.83 & 22.98 & 18.08 & 15.83 & 22.94 & 16.20 & 12.91 & 12.38 & 11.78 & 9.57  & 7.67 & 7.14 & 6.50 & 6.12 \\ \midrule
Uniform & 4/16 & 34.22 & 27.41 & 20.56 & 17.52 & 22.98 & 16.20 & 12.95 & 12.38 & 11.78
 & 10.25 & 9.12 & 7.94 & 7.09 & 6.83 \\
Non uniform & 4/16 & 31.84 & 25.39 & 19.55 & 16.75 & 22.98 & 16.20 & 12.91 & 12.38 & 11.78 & 10.21 & 8.23 & 7.56 & 6.79 & 6.34\\ \midrule
Uniform & 8/8 & 34.84 & 35.81 & 21.59 & 20.64 & 26.88 & 17.94 & 98.88 & 6116& 2790& 4931 & 20.05 & 18.00 & 19.36 & 18.22 \\
Non uniform & 8/8 & 28.33 & 23.08 & 18.14 & 15.89 & 22.98 & 16.42 & 12.95 & 12.40 & 11.78 & 10.21 & 7.68 & 7.16 & 6.52 & 6.13 \\ \midrule
Uniform & 4/8 & 43.44 & 45.19 & 25.09 & 22.8 & 26.92 & 17.86 & 100.81 & 6116& 2812 & 5191 & 39.09 & 28.94 & 30.39 & 28.05 \\
Non uniform & 4/8 & 32.53 & 25.48 & 19.62 & 16.81 & 22.98 & 16.42 & 12.95 & 12.38 & 11.78 & 10.34 & 8.26 & 7.58 & 6.81 & 6.38 \\ \midrule
\multicolumn{2}{l}{\textit{Lambada (accuracy$\uparrow$)}}
\\ 
\midrule
full & 16/16 & 64.59 & 69.40.& 73.98 & 77.39 & 67.57 & 75.35 & 81.25 & 80.71 & 81.99 & 83.01  & 88.58 & 89.42 & 90.31 & 90.82 \\ \midrule
Uniform & 4/16 & 58.08 & 65.00 & 72.85 & 76.69 & 63.50 & 59.01 & 75.19 & 77.28 & 69.89 & 41.01  & 85.77 & 87.06 & 88.09 & 88.77 \\
Non uniform & 4/16 & 61.88 & 68.27 & 71.68 & 76.96 & 64.37 & 70.79 & 79.22 & 79.44 & 80.55 & 81.63 & 87.08 & 88.64 & 89.98 & 90.29 \\ \midrule
Uniform & 8/8 & 59.77 & 61.90 & 68.62 & 66.87 & 64.61 & 72.91 & 41.75 & 5.38 & 3.37 & 0.84 & 86.03 & 84.10 & 65.00 & 80.45 \\
Non uniform & 8/8 & 64.70 & 68.99 & 73.77 & 77.39 & 67.65 & 74.76 & 80.94 & 80.47 & 81.84 & 82.49 & 88.52 & 89.46 & 90.37 & 90.32\\ 
\bottomrule
\end{tabular}}
\end{center}
\caption{Comparison of uniform and non-uniform method on three datasets. In addition to evaluating W4A16 and W8A8 quantization, we also evaluate the much more challenging w4a8 quantization setting where weights are quantized to 4 bits and activations to 8 bits. For this experiment, we have not processed the entire test set for Wikitext-2 and C4. Instead, we use the first 2000 samples to calculate perplexity.}
\label{tab:1}
\end{table*}

To discuss the impact of extreme perturbations caused by clipping, we design and experiment with this special perturbation:
\begin{itemize}
    \item \textbf{Clipping Perturbation} $\Delta_{\text{C}}$: where large values exceeding the pre-set threshold are clipped to the threshold. We set the clipping threshold as $\mu_{\textbf{X}}+k\cdot\sigma_{\textbf{X}}$.
\end{itemize}

Figure~\ref{fig:01} illustrates the effect of clipping perturbations on performance. Overall, clipping has a considerably more adverse impact than regular perturbations. This finding aligns with the outlier phenomenon, which posits that extremely large values are crucial for LLM performance. Clipping these values results in significant information loss, thus leading to severe performance degradation.
Another interesting observation is the comparison between LLAMA and BLOOM, where models from the LLAMA family seem to be more vulnerable to clipping.

\begin{figure}[t]
  \centering
  \includegraphics[width=\linewidth]{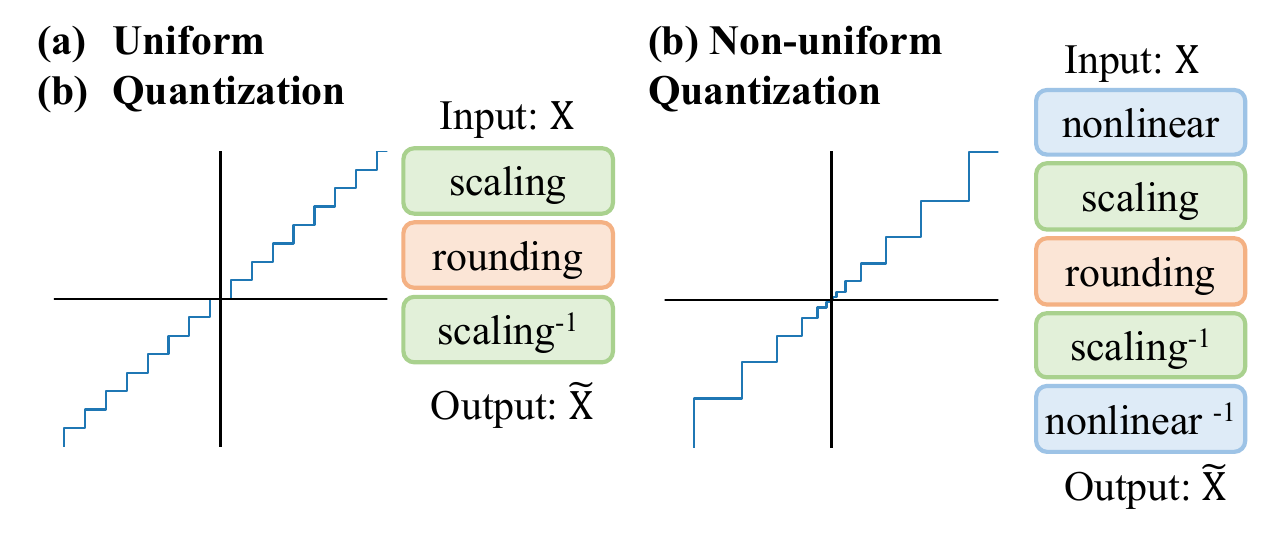}
  \caption{Illustration of uniform quantization and its non-uniform adaptation. The key difference is that non-uniform quantization utilizes two mutually inverse transformations to shape the input in a more quantization-friendly manner, one before the scaling operation and the other after the scaling-back operation.}
  \label{fig:4}
\end{figure}

\begin{table}
\small
    \centering
    \scalebox{1}{\begin{tabular}{llcc}
    \toprule
        Quant. Methods & Bits & Wikitext-2 & C4  \\ \midrule
        Full Precision& 16/16 & 19.70 & 7.67\\ \midrule  
        Non-uniform (Ours) & 4/16 & 20.92 & 8.23\\ 
        RTN (alias of Uniform Quant.) & 4/16 & 25.05 & 9.12\\
        GPTQ~\cite{frantar2022gptq} & 4/16 & 21.09  & 8.48\\
        AWQ~\cite{lin2023awq} & 4/16 & 20.77 & 8.03\\

    \bottomrule
            
    \end{tabular}}
    \caption{Comparison of different methods on LLaMA-7b.}
    \label{tab:x}
\end{table}

\section{Improving LLM Quantization}
Having identified the preferred types of perturbations, we can now explore how these properties can be leveraged to improve existing quantization methods. Specifically, we have discovered that smaller values in a tensor should be subject to smaller perturbations, while larger values can handle greater perturbations, and that the magnitude of the perturbations should be constrained and clipping should be avoided. These properties can be useful in designing a quantization method that takes them into account.

To this end, we have implemented a quantization method to verify the effectiveness of the properties we discovered. Our goal is to ensure that smaller values in a tensor are subject to smaller perturbations, while larger values bear more perturbations. We achieve this by applying a nonlinear transformation before quantization, which amplifies smaller values around zero and shrinks larger values. The approach is straightforward: perform a nonlinear transformation before quantization, and inverse transform it back after quantization, as illustrated in Figure~\ref{fig:4}. For simplicity, we select $f(x)=x^{1/3}$ as the nonlinear transformation as it effectively amplifies values near zero and shrinks larger values. 

\begin{table}
\small
\setlength\tabcolsep{4.pt}
    \centering
      \scalebox{1}{\begin{tabular}{llllll}
        \toprule
        Method & Humanities &STEM & Social  & Other & Avg.  \\
        &&&Sciences \\
        \toprule
        Full Prec. & 34.80 & 35.12 & 30.29 & 38.07 & 37.95 \\
        Uniform & 31.63 & 34.13 & 27.55 & 33.38 & 33.92 \\
        Non-uniform & 34.43 & 33.29 & 31.48 & 38.40 & 35.83 \\
        
        \bottomrule
      \end{tabular}}
      \caption{LLaMA-7B 4-shot ICL results on MMLU.}
      \label{tab:y}
\end{table}

We conducted experiments on three model families: BLOOM, OPT, and LLAMA.  Table~\ref{tab:1} summarizes results on Lambada, Wikitext-2, and C4. In the W4A16 setting, where activations are not quantized, uniform quantization performs well on most of the models and tasks, suggesting that weight quantization is not overly challenging. However, when it comes to quantizing activations, uniform quantization exhibits weaknesses, while the non-uniform method maintains near-lossless performance. We observe that for OPT models, as the model scales up, standard min-max uniform quantization struggles to perform accurately with explosive perplexity scores, whereas the non-uniform method maintains the model's performance close to that of the unquantized model. Another interesting finding is that for non-uniform quantization, as the model size scales up, the performance gap between before and after quantization decreases, indicating that larger models are easier to quantize. This phenomenon contrasts with observations made in uniform quantization, where model performance deteriorates as the model size grows.

We conducted more experiments on LLaMA-7b. Table~\ref{tab:x} shows the comparison of non-uniform quantization and some recent post-training quantization methods and Table~\ref{tab:y} is the performance on MMLU~\cite{DBLP:conf/iclr/HendrycksBBZMSS21}, an advanced benchmark for evaluating the abilities of LLMs. 

\paragraph{Limitations and Practical Value}
Quantization offers two sources of efficiency improvements: reduced memory costs through the use of low-precision formats, which accelerates data transfer in Memory/GPUs (memory efficiency), and faster matrix multiplication of low-precision formats (computation efficiency). However, the latter advantage is only realized when both the weight and activation are quantized to low precision. In the commonly used W4A16 setting, matrix calculation is still performed with high precision.
Despite non-uniform quantization being a dominant approach in signal processing, it has not received sufficient attention in the realm of neural network compression. One major obstacle to its application in model compression is efficiency concerns. Although our non-uniform quantization method demonstrates reliable performance, it is important to note that the introduction of nonlinear transformation entails additional computations. Consequently, non-uniform quantization only offers memory efficiency without computation efficiency.
However, as current LLMs are not computation-constrained but rather memory-constrained, the model wastes a significant amount of time waiting for variables to be loaded on specific hardware rather than performing calculations. This imbalance of memory capacity and computation is particularly outstanding when generating long text, as the loading/offloading of KV-Cache for each generated token significantly slows down inference. This scenario is where most of the current quantization approaches glow.

\paragraph{Further Insights}
Through our experimentation with the non-uniform quantization method, which amplifies smaller values and compresses larger ones, we have confirmed a general principle regarding model quantization. Specifically, larger values can tolerate more error, while smaller values are more sensitive to larger errors. Based on this observation, we have identified the limitations of uniform quantization, which employs uniformly distributed quantization intervals to scale all values of varying sizes. We propose that is preferable to establish dense bins for small values and sparser bins for large values.
We believe that this finding is highly thought-provoking and warrants further exploration in future research on model quantization. There are two potential avenues to pursue: one is to implement hardware-friendly non-uniform quantization for efficient deployment, while the other entails shaping the weight and activation distributions through training to make them easier to quantize. Some prior research has already been conducted in this direction~\cite{bondarenko2023quantizable}.


\section{Conclusion}
Our work introduces a new perspective on quantization, which we refer to as ``the lens of perturbation". Using this approach, we conduct a comprehensive investigation of uniform quantization on LLMs, evaluating the performance of various models under different quantization settings. We also probe the models with manually constructed perturbations, which provide valuable insights into how quantization impacts the model performance. These insights help us understand the difficulties associated with quantization for LLMs. Based on our findings, we propose a non-uniform quantization approach that significantly reduces the performance degradation caused by quantization. We hope that our study, conducted through the lens of perturbation, will contribute to a better understanding of the challenges associated with quantization for LLMs and inspire the development of more efficient and effective quantization methods.

\section*{Acknowledgments}

This work is supported by National Key R\&D Program of China (No. 2022YFC3301900) and National Natural Science Foundation of China (NSFC Grant No. 62122089). We sincerely thank all reviewers for their valuable comments and suggestions, which are crucial for improving our work.
\appendix

\bibliography{aaai24}

\end{document}